\documentclass[10pt,a4paper,twocolumn]{article}

\usepackage{graphicx}
\usepackage{hyperref}
\usepackage[utf8]{inputenc}

\author {
  Thomas Haschka$^1$ \\ thomas.haschka@tuwien.ac.at  \and Joseph Bakarji$^2$ \\ jb50@aub.edu.lb 
  }

\date{
$^1$ E020-04 Service Unit of High Performance Computing, DataLab,
Campus IT, \\ Technische Universität Wien, Operngasse 11, Vienna, 1040,
Vienna, Austria \\
$^2$ Department of Mechanical Engineering, American University of Beirut,\\
P.O. Box 11-0236, Riad El-Solh, Beirut, 1107 2020, Beirut, Lebanon. \\[2ex]
\today
}

\title{Discovering Multi-Scale Semantic Structure in Text Corpora Using Density-Based Trees and LLM Embeddings}

\begin{document}
\maketitle

\section*{Abstract}

Recent advances in large language models enable documents to be
represented as dense semantic embeddings, supporting similarity-based
operations over large text collections. However, many web-scale
systems still rely on flat clustering or predefined taxonomies,
limiting insight into hierarchical topic relationships. 
In this paper we operationalize hierarchical density modeling on large
language model embeddings in a way not previously explored. Instead of
enforcing a fixed taxonomy or single clustering resolution, the method
progressively relaxes local density constraints, revealing how compact
semantic groups merge into broader thematic regions. The resulting
tree encodes multi-scale semantic organization directly from data,
making structural relationships between topics explicit. 
We evaluate the hierarchies on standard text benchmarks, showing that
semantic alignment peaks at intermediate density levels and that
abrupt transitions correspond to meaningful changes in semantic
resolution. Beyond benchmarks, the approach is applied to large
institutional and scientific corpora, exposing dominant fields,
cross-disciplinary proximities, and emerging thematic clusters. 
By framing hierarchical structure as an emergent property of density
in embedding spaces, this method provides an interpretable,
multi-scale representation of semantic structure suitable for large,
evolving text collections.

\section{Introduction}

Understanding the semantic organization of large text corpora is a
central challenge in knowledge representation, information retrieval,
and semantic web research. As digital collections continue to grow in
size and diversity, there is an increasing need for methods that not
only group documents by similarity but also expose how concepts relate
to one another across multiple levels of abstraction. Such structure
is essential for tasks ranging from exploratory search and corpus
navigation to the analysis of disciplinary landscapes and knowledge
evolution. 

Recent advances in large language models (LLMs) have substantially
improved the representation of textual meaning. High-dimensional
embedding models encode semantic similarity with remarkable fidelity
and have become a standard component of modern information systems,
particularly in vector databases and semantic retrieval
pipelines. While these embeddings enable effective similarity search
and flat clustering, they typically provide limited insight into the
global organization of semantic space. In particular, they do not
directly reveal how fine-grained concepts combine into broader themes,
nor how semantic resolution changes as one moves from local to global
structure. 

Hierarchical representations offer a natural solution to this
limitation. By organizing documents into nested structures,
hierarchies can capture relationships between topics at multiple
semantic scales. Traditional hierarchical approaches in text analysis,
however, often rely on strong modeling assumptions, predefined
taxonomies, or global distance matrices that are sensitive to noise
and high dimensionality. As a result, they can struggle to scale to
large corpora or to adapt flexibly to heterogeneous textual content. 

In this work, we propose a data-driven approach for reconstructing
semantic hierarchies directly from text embedding spaces. Rather than
imposing a predefined ontology or topic structure, we infer in an
unsupervised manner hierarchical relationships by progressively
revealing structure in the embedding space itself. The core idea is to
identify groups of highly similar, documents and to examine how these
groups merge as semantic similarity criteria are gradually
relaxed. This process yields a tree-like structure in which terminal
branches correspond to tightly related (dense) texts, while higher-level nodes
represent increasingly (diffuse) general semantic groupings. 

A key advantage of this approach is that it produces a single, unified
representation that simultaneously captures local semantic coherence
and global thematic organization. The resulting trees are not
restricted to a fixed number of clusters or levels and naturally
accommodate documents that do not belong to any dense semantic group
at finer resolutions. Such behavior is particularly important for
real-world corpora, where topical boundaries are often fuzzy and
unevenly populated. 

We demonstrate the utility of this framework on several benchmark
datasets commonly used in text classification and topic analysis,
including 20 Newsgroups, IMDB 50K reviews, and AG News. These
experiments allow us to assess how well the inferred hierarchies align
with established labels and to analyze how semantic agreement varies
across hierarchical depth. Beyond benchmarks, we apply the method to
large collections of scientific abstracts from academic institutions,
where no single ground-truth taxonomy is assumed. In these cases, the
inferred trees reveal institution-specific research profiles,
cross-disciplinary connections, and emerging thematic areas. 

To support interpretability, we further integrate large language
models as semantic annotators. By summarizing the textual content
associated with individual tree nodes, we generate concise,
human-readable descriptions of semantic regions within the
hierarchy. This results in an end-to-end pipeline that combines
embedding-based structure discovery with automated semantic labeling. 

Overall, this work contributes a flexible and interpretable method for
uncovering hierarchical semantic structure in large text corpora. By
emphasizing multi-scale organization, data-driven discovery, and
semantic transparency, the proposed approach aligns naturally with the
goals of semantic web research and provides new tools for exploring
complex textual knowledge spaces. 

\section{Related Work}

\subsection{Text Representation for Semantic Analysis}

Text classification and organization have long relied on vector-based
representations to capture semantic relationships within document
collections. Early approaches such as bag-of-words models represent
texts as unordered collections of terms, enabling efficient comparison
but ignoring syntactic structure and semantic context
\cite{bagofwords}. Subsequent developments introduced weighting
schemes such as term frequency–inverse document frequency (tf–idf),
which improved discrimination by accounting for term importance within
a corpus \cite{tfidf}. 

Latent Semantic Analysis (LSA) extended these ideas by projecting
document-term matrices into lower-dimensional latent spaces using
singular value decomposition, enabling the discovery of latent
semantic associations \cite{lsa-first,lsa,lsa-intro}. Probabilistic
topic models, most notably Latent Dirichlet Allocation (LDA) and its
hierarchical extensions, introduced generative frameworks for
uncovering thematic structure in document collections
\cite{lda,hda}. While effective for identifying coarse-grained topics,
these models require predefined assumptions about topic structure and
often struggle with interpretability at fine semantic resolution
\cite{interpretability}.

Neural embedding models, including word2vec and related architectures,
marked a transition toward distributed semantic representations
learned from large corpora \cite{word2vec}. However, these approaches
primarily capture word-level proximity and do not fully encode
compositional or contextual meaning. The introduction of
transformer-based architectures \cite{transformer} and large language
models (LLMs) enabled the generation of contextualized document-level
embeddings that capture syntactic structure, semantics, and
discourse-level information.

LLM embeddings are now widely used in information retrieval, semantic
search, and retrieval-augmented generation pipelines
\cite{rag,llm-embed-rag}. Despite their expressive power
\cite{embeddings-proof,embeddings-proof-more,embedding-proof-even-more}
their use in information organization is often limited to nearest-neighbor
retrieval or flat clustering, leaving the global semantic structure of
document collections largely unexplored.

\subsection{Clustering and Hierarchical Organization of Text}

Beyond representation, organizing documents into meaningful structures
is a core challenge in information processing. Flat clustering methods
such as k-means \cite{k-means-first,k-means} or DBSCAN \cite{dbscan}
are frequently applied to document embeddings to identify groups of
related texts. While computationally efficient, flat partitions impose
a single resolution of analysis and obscure relationships between
topics at different levels of semantic abstraction.

Hierarchical approaches aim to overcome this limitation by explicitly
modeling multi-level structure. Agglomerative hierarchical clustering
constructs dendrograms from pairwise distances but requires global
distance matrices and linkage criteria that are sensitive to noise,
scaling, and high dimensionality. Similarly, hierarchical topic models
rely on strong probabilistic assumptions and often exhibit limited
robustness when applied to large, heterogeneous corpora
\cite{hierarchical-first,hierarchical}.

Density-based clustering offers an alternative by identifying regions
of high local similarity without assuming convex cluster
shapes. While such methods
have been successfully applied in spatial data analysis and other
high-dimensional domains \cite{mnhn-tree-tools,md-tree-tools},
their use in text and information retrieval
has largely focused on extracting flat clusterings rather than
interpreting the full hierarchical structure. As a result, the
potential of density-based hierarchies for revealing multi-scale
semantic organization in text corpora remains to be explored. 

\subsection{From Density-Based Clustering to Semantic Trees}

Hierarchical density-based clustering has been successfully applied in
other high-dimensional domains, such as bioinformatics and molecular
dynamics, to reconstruct phylogenetic relationships and
conformational landscapes \cite{mnhn-tree-tools, md-tree-tools}. In
these settings, trees are constructed by progressively relaxing
density constraints, embedding compact structures into broader ones
and yielding interpretable representations of multi-scale structure.

This work adapts this paradigm to the organization of document
embeddings. Rather than treating the hierarchy as an intermediate
artifact or selecting a single optimal partition, we construct and
analyze the full density-based tree as the primary representation of
semantic structure. The resulting semantic tree captures how documents
group and merge across similarity scales, enabling analysis at multiple
levels of abstraction, from fine-grained topical distinctions to broad
thematic domains.

Crucially, the construction is independent of predefined taxonomies or
label sets. This makes the approach well suited to exploratory analysis
of large and heterogeneous corpora, where existing classifications are
often incomplete, inconsistent, or entirely absent.

\subsection{Interpretability and Automatic Annotation}

Hierarchical organization alone is insufficient without
interpretability. Large trees derived from embedding spaces require
semantic summaries that allow users to understand and navigate the
structure. Recent advances in LLMs enable large-context inference,
making it possible to generate concise semantic descriptions for
collections of documents. 

We leverage this capability to automatically annotate tree nodes by
prompting LLMs with the textual content of each cluster. This produces
human-readable labels that summarize the dominant themes represented
at different levels of the hierarchy. The result is an end-to-end
pipeline that combines unsupervised structure discovery with automatic
semantic interpretation, addressing a long-standing challenge in
large-scale document organization. 

\section{Methodology}

\subsection{Datasets} \label{sec-datasets}

The primary goal of this study was to develop a pipeline for the
automated classification of research fields based on textual
similarity. While motivated by this scientific application, the
proposed approach is general and can be applied to a broad range of
text corpora. To ensure linguistic consistency and to avoid potential
errors caused by multilingual embeddings and the tree building
process, which would in such a case probably attributed different
branches to different languages, we restricted our analysis
to texts written in English.

We selected the following datasets for our study:
\begin{itemize}
  \item \textbf{AUB}: English abstracts of publications authored by the
    American University of Beirut (AUB) from 2018 to October 2025. The data
    were provided by the AUB University Libraries. To ensure comparability,
    the roughly the same temporal window and language filters, only english
    abstracts, as for the TU Wien corpus were applied.
  \item \textbf{TU Wien}:   English abstracts of publications affiliated
    with TU Wien between 2018 and April 2025. The data were provided by
    the TU Wien Service Unit for Research Information Systems as
    aggregated extracts from the institutional repository
    (\emph{reposiTUm}), Scopus, and Dimensions databases. Duplicates
    were removed, and only records containing English abstracts were
    retained. 
  \item \textbf{20 Newsgroups}: The 20 Newsgroups dataset is a widely
    used benchmark corpus for 
    text classification and topic modeling. It consists of
    approximately 20,000 newsgroup posts collected from 20 distinct
    discussion groups on topics such as science, politics, sports,
    religion, and technology \cite{20newsgroups}.
  \item \textbf{IMDB 50K Reviews}:
     A corpus of 50,000 movie reviews from
     the Internet Movie Database (IMDB), each labeled with a binary
     sentiment (positive or negative) \cite{50kmovies}. 
   \item \textbf{AG News}: A large-scale news dataset consisting of short
     article titles and descriptions categorized into four topics: World,
     Sports, Business, and Science/Technology.
     \cite{agnews}.
   \item \textbf{Theses-fr-en}: English abstracts of theses defended in
    France: The French
    government publishes by means of its open data initiative a
    dataset containing all theses defended in France since 1985. We
    filtered this dataset for english abstracts. This data was
    obtained from the \emph{data.gouv.fr} platform.
\end{itemize}
Statistics about the different datasets are shown in table
\ref{tab-dataset-statistics}.

\begin{table}
  \caption{Statistics about the text corpa used. \emph{n(Texts)} is
    the number of unique texts in the dataset, \emph{mean} and
    \emph{stddev} represent the mean and standard deviation of the
    number of words contained in each text of the dataset.}
  \label{tab-dataset-statistics}
  \begin{tabular}{l|rrr}
    \hline
    & n(Texts) & mean & stddev \\
    \hline
    AUB & 15982 & 237.73 & 98.24 \\
    TU Wien & 27506 & 168.85 & 73.79 \\
    20newsgroups & 11314 & 287.47 & 541.46 \\
    50k-movies & 50000 & 231.16 & 171.34 \\
    ag\_news & 127000 & 37.84 & 10.09 \\
    theses-fr-en & 180939 & 261.89 & 112.70 \\
    \hline
  \end{tabular}
\end{table}
    
\subsection{Embeddings}
Embeddings for the datasets outlined in section \ref{sec-datasets}
were generated using either the Qwen3-Embedding-8B
\cite{qwen3embedding} or SFR-Embedding-Mistral \cite{SFRembedding}
model. These models were chosen selected based on their high performance in the
MTEB-Leaderboard found on Huggingface, and their practical
deployability on the computing infrastructure accessible to us.

4096-dimensonal embedding vectors obtained for each text from a
dataset were stored in an in house developed vector database.

Each text sample was transformed into a 4096-dimensional embedding
vector. The resulting embeddings were stored in an in-house developed
vector database optimized for high-throughput semantic similarity
retrieval.

\subsection{Tree Building}

\subsubsection{Pairwise Distance and Dimensionality Reduction}

The pairwise distance between embeddings serves as the foundation for
the hierarchical tree construction process. We employed the cosine
distance on the full-length embedding vectors as well as the $L_2$-norm
on reduced-dimensional representations obtained via Principal
Component Analysis (PCA). 

Dimensionality reduction proved crucial, as PCA functions as a feature
selection mechanism that prioritizes components with maximum variance.
Empirically, we observed that reducing the number of components leads
to more clearly separated and interpretable cluster structures. While
the original embedding vectors contain 4096 dimensions, spectral
analysis indicated significant redundancy among components. The choice
of the number of principal components thus directly affects the
resolution and spread of the resulting semantic tree.

\subsubsection{Adaptive Density Clustering with DBSCAN}

Semantic clustering of text documents was performed using a
density-based method adapted from DBSCAN \cite{dbscan}. At iteration
$L$, clusters were defined as regions where the local density $\rho_L$
satisfies 
\begin{equation}
\rho_L > \frac{\mathrm{minpts}}{V(\epsilon_L)},
\end{equation}
where $\epsilon_L$ is the neighborhood radius defining a volume
$V(\epsilon_L)$, and $\mathrm{minpts}$ is the minimum number of points
required within that volume. Unlike HDBSCAN \cite{hdbscan}, which
constructs a hierarchy internally but typically outputs a single flat
clustering, this method explicitly preserves hierarchical structure
across successive density thresholds. 

To capture semantic structure at multiple scales, $\epsilon_L$ was
increased in fixed increments $\Delta\epsilon$, generating
progressively lower density thresholds. Clusters identified at each
step contained denser clusters from previous iterations. For iteration
$L+1$, the density criterion becomes 
\begin{equation}
\rho_{L+1} > \frac{\mathrm{minpts}}{V(\epsilon_L + \Delta\epsilon)}.
\end{equation}

\begin{figure*}[!htb]
  \includegraphics{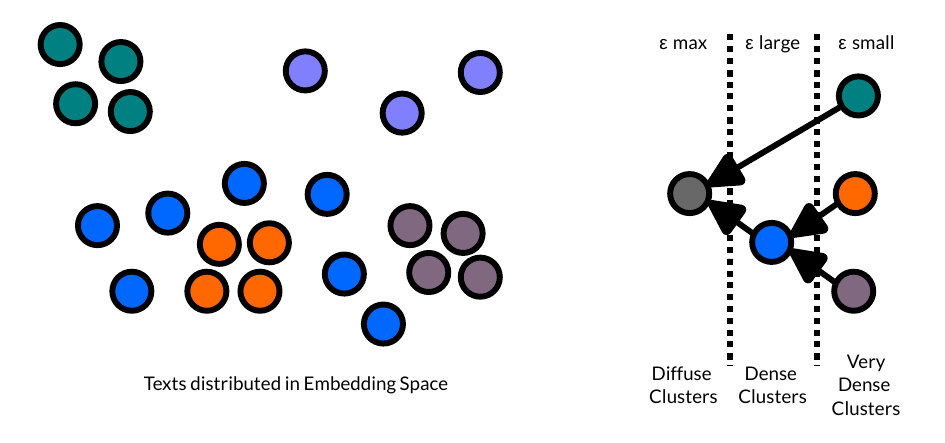}
  \caption{Tree construction from a dataset represented in embedding
    space. Left: embedding space with clusters of different densities;
    each colored sphere represents a text document. Right: conceptual
    tree built from the embedding space, showing how denser clusters
    (orange and grape) merge into larger ensembles (blue) as the
    density threshold is relaxed, while other clusters (green) remain
    separate and some points (violet) are considered
    outliers. Eventually, all points merge into a single root cluster
    (gray). }
  \label{fig-tree-from-space}
\end{figure*}

For each dataset, hierarchical trees were constructed using successive
runs of DBSCAN. Each iteration is controlled by three parameters: the
initial radius $\epsilon_0$, the incremental step $\Delta\epsilon$,
and the minimum points $\mathrm{minpts}$. In each successive run, the
radius is increased while keeping $\mathrm{minpts}$ constant,
progressively lowering the effective density threshold. Previously
dense clusters merge into larger, more diffuse clusters, and whenever
fewer clusters are found than in the preceding iteration, the new
clusters are stored, embedding the previous ones within
them. Figure~\ref{fig-tree-from-space} illustrates this process
conceptually. 

Unlike methods that optimize a predefined objective (e.g., topic
coherence, likelihood, or cluster compactness), this approach
constructs a hierarchical semantic representation directly from local
density relations in the embedding space. It allows semantic groupings
to emerge across multiple resolutions without assuming a fixed
taxonomy, linkage criterion, or global distance structure. To our
knowledge, this is the first method that operationalizes nested
density relaxation as a general-purpose mechanism for semantic tree
induction in text corpora. 

\subsection{Tree Annotation}

The resulting trees consist of multiple hierarchical layers. Each
layer represents a set of clusters that form the nodes at that level,
while edges connect clusters of higher density (outer layers) to
broader, lower-density clusters (inner layers). This structure allows
fine-grained semantic groups to merge progressively into coarser
domains.

To realize our goal of a data-driven research field classification
pipeline, we sought to automatically annotate each node with a
representative topic label. The underlying assumption is that smaller,
high-density clusters correspond to specialized research areas, while
broader clusters capture more general scientific domains.

To generate these annotations, we employed a Large Language Model
(LLM) capable of handling large textual contexts. Specifically, we
used the Llama-4-Scout-17B-16E-Instruct model
\cite{meta2024llama4} in 6-bit quantized form,
which provides a nominal context window of up to
11 million tokens. Due to technical constraints, our inference
pipeline successfully operated with a maximum context size of
approximately 600,000 tokens.

Both the TU Wien and AUB corpora were fully annotated, encompassing
all English-language scientific abstracts published since 2018. For
each node, the LLM received the text of all abstracts within that
cluster and was prompted to infer a unifying research field label. 

In cases where a cluster contained more than 1500 abstracts (exceeding
the 600,000-token limit), we randomly sampled 1500 abstracts per
processing chunk. For $n$ total abstracts in a cluster, the number of
chunks $s$ was determined as:$
  s = \frac{3n}{1500}$,
where the factor of 3 ensures oversampling to enhance topic stability.
Each chunk was processed independently by the LLM to yield a candidate
research field label. In less than 1 in 100 cases, the LLM did not
produce a clear output. Even though the prompt enforces to yield only
one research field on a single line, this is not always the case. As
such in all cases where answers span more than one line, or contain
more than 80 characters, we
yield these answers to the gpt-oss-20b model \cite{gpt-oss} in 8-bit
quantized form, and ask it
to decide from the previous output on a single research field. As the
output from the first run might be due to errors in the LLM very large
we only keep the last 20000 characters during this pass to the
gpt-oss-20b model. 

\begin{figure}[!htb]
  \begin{center}
    \includegraphics{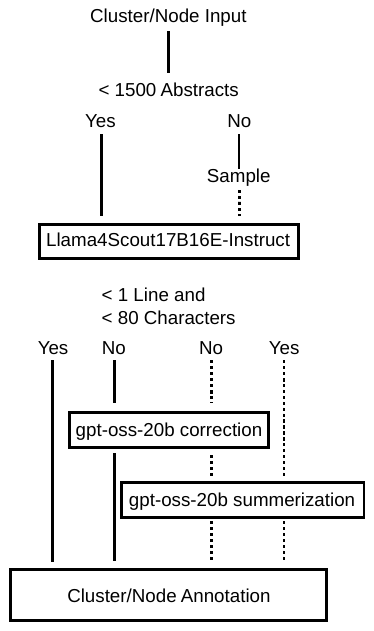}
    \caption{Schematic representation of the tree node / cluster
      annotation process.}
    \label{fig-annotation}
  \end{center}
\end{figure}

In the case where the cluster had to be sampled because it contained
more than 1500 abstracts all resulting labels were subsequently fed into
a second LLM pass. Again the gpt-oss-20b model \cite{gpt-oss} in 8-bit
quantized form is used in such a case where the model inferred the
overarching research field encompassing all $s$ subfields. An overview about
the different paths that annotation might undergo is outlined in Figure
\ref{fig-annotation}.

Tree annotation is computationally intensive. The AUB tree required
annotating 33{,}862 text files, while the TU Wien tree required
62{,}012, including oversampling. Inference was performed using the
\texttt{llama.cpp} \cite{llama-cpp} engine on the MUSICA supercomputer
(Austria). Up to 80 NVIDIA H100 GPUs were employed concurrently during
annotation. Each compute node with four H100 GPUs (80~GB VRAM each)
handled batches of approximately 600{,}000 tokens per inference
instance. Multi-node inference, which could allow for a higher context
token limit, was not attempted as MUSICA is currently in its test
phase; instead, parallelism was achieved by distributing text files
across multiple nodes for simultaneous inference.

\subsection{Evaluation Metrics for Hierarchical, non-Consistent Partitions}

To quantitatively assess how well the inferred semantic trees align
with existing labels in benchmark datasets, we employ a combination of
external clustering validation measures: Adjusted Rand Index (ARI) and
Normalized Mutual Information (NMI). For the evalution of these
metrics we make use of their implementation in Scikit-Learn
\cite{scikit-learn}. These metrics
are applied independently at each tree layer, treating the clustering
at that layer as a flat partition of the dataset. 

ARI and NMI are widely used measures for comparing a predicted
clustering against a ground-truth labeling. ARI evaluates agreement
between two partitions based on pairwise co-assignment, adjusted for
chance, while NMI measures the shared information between cluster
assignments and class labels, normalized to account for differing
numbers of clusters. In our setting, these metrics are computed per
tree layer using only the documents assigned to clusters at that
layer. This is necessary because the density-based procedure may leave
a fraction of documents unclustered at higher density levels, leading
to non-exhaustive and non-consistent partitions across layers. While
ARI and NMI are not inherently designed for hierarchical or incomplete
clusterings, they remain informative for assessing local alignment
between inferred clusters and known labels where such overlap exists.

\subsection{Label Transcription} \label{sec-transcription}

The theses-fr-en dataset is annotated and contains a research field
(discipline, in French) for each abstract. However, these fields can
in many cases be freely chosen by the thesis author, resulting in 13,599
unique discipline labels.
In a first test we cross annotated these unique discipline labels into
a binary classification of either humanities or natural sciences. We
then proceeded to obtain a more coherent and standardized annotation,
we used the gpt-oss-20b model to classify all 13,599 labels according to the
OECD Fields of Science (FOS) 2002 classification \cite{oecd-fos}. 
The resulting transcribed labels were then used to color
the trees inferred from this dataset, allowing us to assess whether
specific scientific fields tend to separate into distinct
and meaningful branches. 

\subsection{Tree Coloration and Visualization}

To facilitate interpretation and qualitative validation of the inferred
trees, we applied coloration schemes that project existing labels or
metadata onto the tree structure. This process enables us to visualize
how known categories or attributes distribute across different branches
of the tree.

For example, in the case of the 20~Newsgroups dataset, tree
coloration reveals how posts from individual newsgroups are positioned
within the hierarchical semantic structure. Similarly, for the
TU~Wien dataset, coloration based on existing research
classifications allowed us to assess how well our hierarchical density
clustering recovers semantically coherent domains.

Tree coloration was performed using utilities provided by
MNHN-Tree-Tools \cite{mnhn-tree-tools}. In particular, we
employed the \texttt{tree\_map\_for\_split\_set} tool to map predefined
clusters or labeled subsets onto the corresponding nodes of the
hierarchical tree. The resulting color maps were generated for
visualization using the newick-utilities package
\cite{newick-utilities}, a software suite for the manipulation and
visualization of phylogenetic trees.

This approach allowed us to produce colored tree visualizations for
each dataset. In the 20 Newsgroups case, individual trees were
colored according to each newsgroup category, highlighting topic
segregation and overlap. For the IMDB 50K reviews dataset, we colored
the tree by sentiment polarity, enabling a visual inspection of how
positive and negative reviews cluster across different density layers.

For the annotated datasets from TU Wien and AUB, we required a
high-resolution visualization that allows all node-level labels to be
displayed and inspected. The resulting tree images are extremely large,
making standard static rendering impractical. To enable detailed
exploration, we developed a tile-based viewer application with two
coordinated windows. One window displays the full, non-annotated tree,
while the second window shows a zoomed view of a selected region,
including the labels for the nodes currently in focus. Navigation is
performed either by clicking on the full-tree view, which updates the
zoomed window to the selected location, or by dragging within the zoomed
window, which moves a red cursor on the full-tree view to indicate the
corresponding position.

\section{Results}

\subsection{Efficiency of PCA and the $L_2$-norm Compared to Cosine Distance}

Cosine distance is widely used for comparing high-dimensional embedding
vectors and is the default similarity measure in many retrieval-based
applications, including RAG pipelines. For this reason, we initially
expected cosine distance to be the natural choice for the DBSCAN
iterations underlying our tree construction algorithm. However, trees
constructed directly from full 4096-dimensional cosine distances
consistently exhibited limited structural resolution: branches were
poorly separated, fine-grained clusters failed to emerge, and the overall
tree topology did not yield significant insights.

In contrast, applying Principal Component Analysis (PCA) prior to
clustering dramatically improved the quality and expressiveness of the
resulting trees. Experiments with 10, 5, and 2 principal components
showed that lower-dimensional PCA representations consistently produced
more “spread out” trees with clearer branching and more detectable
fine-scale semantic substructure. We suggest that this effect is due
to the inherent ability of PCA to yield principal components that
maximize variance and that this feature selection process has a
positive effect on our clustering algorithm. A similar effect was
noted previously in the algorithmic supplement of
\cite{mnhn-tree-tools}.

\begin{figure*}[!htb]
  \begin{center}
  \includegraphics{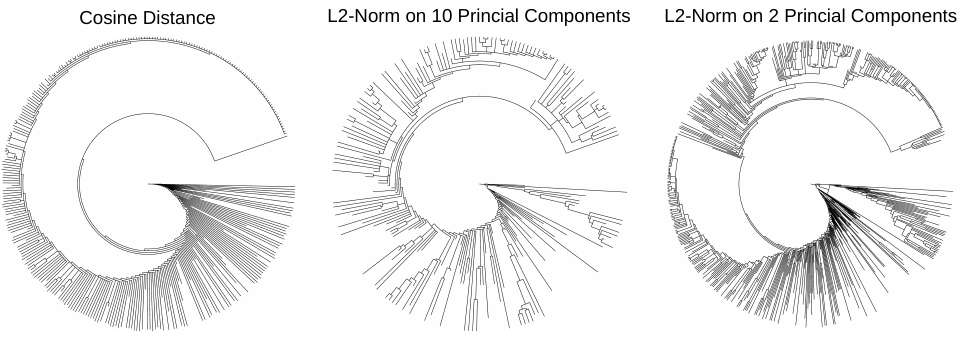}
  \caption{A comparison of trees built from the 20 Newsgroups dataset
    under different distance measures.}
  \label{fig-compairison-L2-cos}
  \end{center}
\end{figure*}

A comparison between trees obtained with cosine distance and with the
$L_2$-norm applied to PCA-reduced embeddings is shown in Figure
\ref{fig-compairison-L2-cos}. From left to right, the trees become more
hierarchically detailed and reveal progressively finer semantic
subclusters. For these illustrative examples, the parameters were chosen
using a rudimentary bisection search to produce visually optimized
trees. The parameter settings were as follows: for cosine distance,
initial $\epsilon = 0.275$, $\Delta \epsilon = 0.001$, and
$\mathrm{minpts}=5$; for the $L_2$-norm on 10 principal components,
initial $\epsilon = 0.091$, $\Delta \epsilon = 0.0001$,
$\mathrm{minpts}=5$; and for the $L_2$-norm on 2 principal components,
initial $\epsilon = 0.0039$, $\Delta \epsilon = 0.00001$,
$\mathrm{minpts}=5$. The trees shown in \ref{fig-compairison-L2-cos}
were built from embeddings obtained from the Qwen3-Embedding-8B model.

\subsection{Detection of Topics and Semantics}

To evaluate whether the proposed hierarchical density-based method
recovers meaningful semantic structure, we combine qualitative analysis
of inferred trees with quantitative evaluation against available ground
truth labels. Since the output of our method is a hierarchy rather than
a single flat partition, evaluation must account for the fact that
clusters emerge, merge, and disappear across tree layers.

For datasets with reference labels, we therefore project the hierarchy
onto a sequence of flat clusterings, one per tree layer, and assess
their agreement with the ground truth using the Adjusted Rand Index
(ARI) and Normalized Mutual Information (NMI). ARI measures the
agreement between two partitions while correcting for chance, whereas
NMI quantifies the amount of shared information between cluster
assignments and labels, normalized to allow comparison across
different numbers of clusters. Both measures are well established in
information retrieval and clustering evaluation and are complementary:
ARI is sensitive to exact cluster membership, while NMI captures
coarser label alignment even when cluster boundaries are imperfect.

Across the benchmark datasets, the evolution of ARI and NMI as a
function of tree depth reveals dataset-specific behavior that reflects
differences in semantic structure rather than methodological
artifacts. For the 20 Newsgroups and AG News corpora, agreement with
ground-truth labels increases from the finest partitions toward
intermediate density levels, reaches a maximum, and then exhibits
sharp decreases followed by extended plateaus as clusters merge at
coarser scales. This pattern indicates that semantic alignment with
predefined categories is strongest at intermediate resolutions, where
fine-grained lexical variation has been absorbed while broader
thematic distinctions are still preserved. Details are outlined in
Figure~\ref{fig-ari-nmi}.

In contrast, the IMDB 50K Reviews dataset exhibits a different
trend. Here, the highest ARI and NMI values occur closer to the leaves
of the tree, within relatively fine-grained branches rather than near
the root as shown in Figure~\ref{fig-ari-nmi}. As density constraints
are relaxed and clusters merge, agreement with 
sentiment labels decreases steadily, and no dominant peak is observed
at intermediate or coarse levels. This behavior reflects the largely
homogeneous semantic structure of the corpus, where sentiment is
encoded as a subtle, distributed signal rather than as a dominant
organizing principle. Consequently, sentiment-aligned clusters emerge
only locally and do not persist as stable large-scale structures. 

Taken together, these results demonstrate that the proposed
hierarchical density-based trees adapt to the intrinsic semantic
organization of each dataset. Rather than enforcing a fixed optimal
resolution, the method exposes multiple meaningful semantic scales,
allowing the analyst to identify the levels at which different types
of structure-topical, thematic, or affective, are most pronounced. 

Importantly, at high-density (outer) layers, only compact and highly
coherent clusters are present, while diffuse documents remain
unassigned. As density constraints are relaxed, clusters merge and
eventually encompass the full dataset. ARI and NMI are therefore
evaluated as a function of tree depth, allowing us to quantify how
semantic agreement evolves across scales rather than at a single,
arbitrarily chosen resolution.

\subsubsection{20 Newsgroups}

\begin{figure*}[!htb]
  \begin{center}
    \includegraphics[scale=0.8]{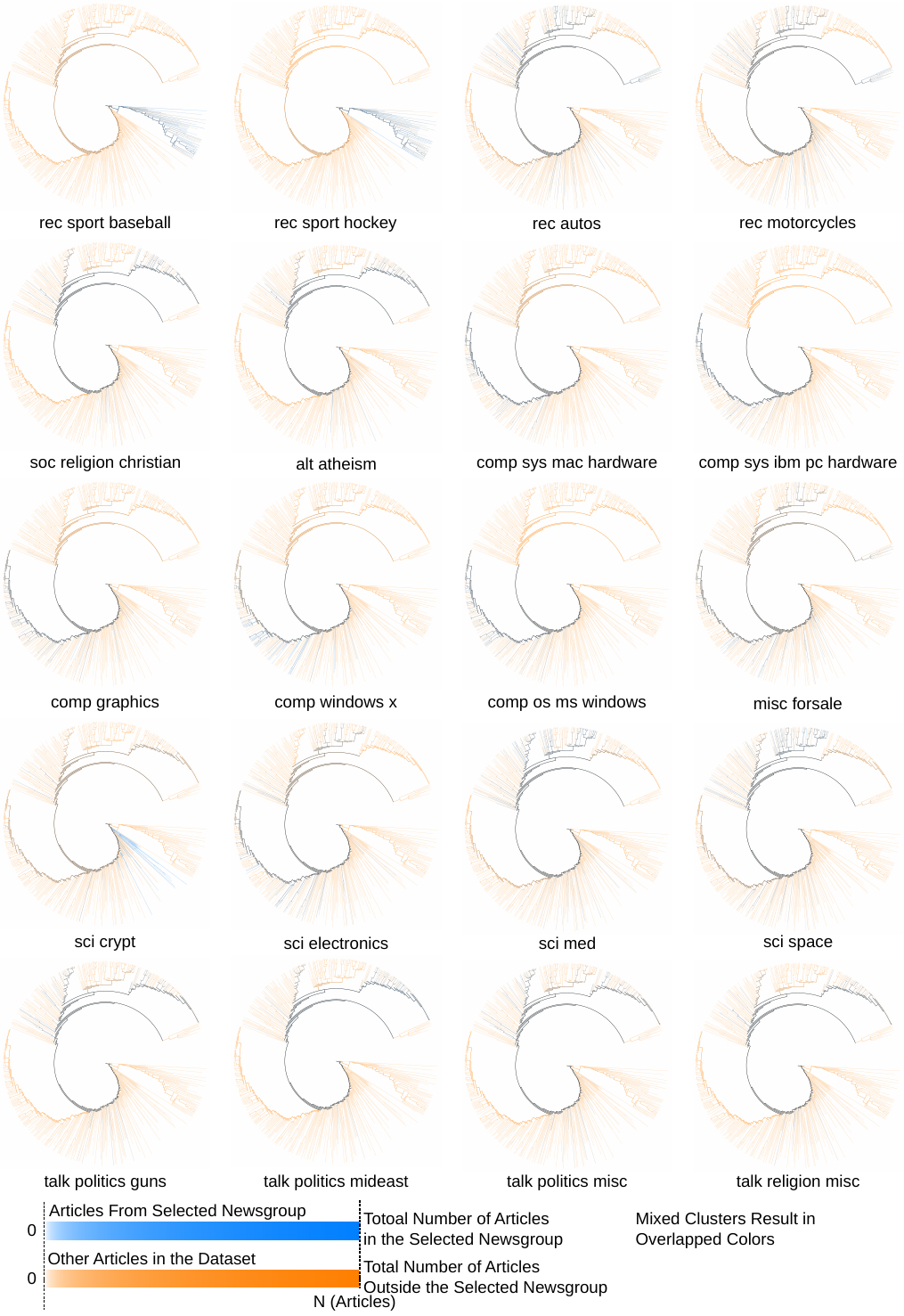}
    \caption{Representations of the tree obtained from the 20
      Newsgroups dataset. Trees were built from the first two
      principal components obtained from Qwen3-Embedding-8B
      \cite{qwen3embedding} embeddings. Parameters used to build the
      tree were: initial $\epsilon = 0.0039$, $\Delta\epsilon = 0.00001$, $\mathrm{minpts} = 5$.}
    \label{fig-20news-colored}
  \end{center}
\end{figure*}

We first apply our method to the 20 Newsgroups dataset, which provides
a well-studied benchmark with twenty topical labels. Figure
\ref{fig-20news-colored} shows the inferred tree, colored according to
individual newsgroup labels.

Quantitatively, ARI and NMI peak at intermediate tree depths,
indicating that the strongest alignment between inferred clusters and
newsgroup labels occurs neither at the finest nor at the coarsest
resolution. At very high densities, clusters are too fragmented to
recover complete topics, while at low densities, semantically distinct
groups merge into broader themes. This behavior is expected for a
hierarchical method and confirms that meaningful topic structure
emerges at specific semantic scales.

Qualitatively, the tree recovers intuitive relationships between
newsgroups:

\begin{itemize}
\item \textbf{Sports-related groups.}  
  The groups \emph{rec.sport.baseball} and \emph{rec.sport.hockey}
  form a shared subtree located close to the root of the full tree.
  This indicates strong internal similarity between the two sports
  discussions, combined with substantial separation from most other
  topics.

\item \textbf{Automotive-related groups.}  
  The groups \emph{rec.autos} and \emph{rec.motorcycles} cluster
  tightly in a small dedicated sub-branch. A fraction of their posts
  appears in regions of the tree that correspond to scientific or
  technical discussions, likely reflecting threads involving
  engineering aspects of vehicles.

\item \textbf{Religion and politics.}  
  The groups \emph{soc.religion.christian}, \emph{alt.atheism}, and
  \emph{talk.religion.misc} form a joint subtree that also contains
  the political groups \emph{talk.politics.guns},
  \emph{talk.politics.mideast}, and \emph{talk.politics.misc}.  This
  intersection reflects the topical proximity and overlapping
  discussions between religion and politics within the dataset.

\item \textbf{Computer-related groups.}  
  The \emph{comp.*} groups, including \newline \emph{comp.sys.mac.hardware},
  \emph{comp.windows.x}, \newline \emph{comp.graphics}, and \newline
  \emph{comp.os.ms-windows}, form a coherent subtree corresponding to
  computer hardware and software discussions. Some overlap with
  \emph{sci.electronics} is visible, consistent with semantically
  similar content.

\item \textbf{Marketplace group.}
  The \emph{misc.forsale} posts distribute across multiple regions of
  the tree, often co-occurring with the automotive and computer
  subtrees, suggesting that sale-related messages tend to mirror the
  underlying topical focus of corresponding technical groups.

\item \textbf{The cryptography group.}  
  The \emph{sci.crypt} group appears in a distinct set of branches
  close to the root, and is clearly separated from both the scientific
  and computer-related subtrees. This separation may stem from the
  unique vocabulary and specialized discourse typical of cryptography
  discussions.
\end{itemize}

These qualitative observations align with the quantitative ARI/NMI
profiles: tree regions that correspond to stable semantic groupings
also contribute most strongly to the metric peaks. Crucially, the tree
representation conveys not only cluster membership but also relative
semantic proximity between topics, which is not captured by flat
clustering alone.

\subsubsection{IMDB 50K Reviews}

We next evaluate the IMDB 50K Reviews dataset, which provides binary
sentiment labels (positive vs. negative). Unlike topical datasets,
this corpus is semantically homogeneous, and sentiment is expressed
largely through subtle linguistic cues.

\begin{figure}[!htb]
  \begin{center}
    \includegraphics{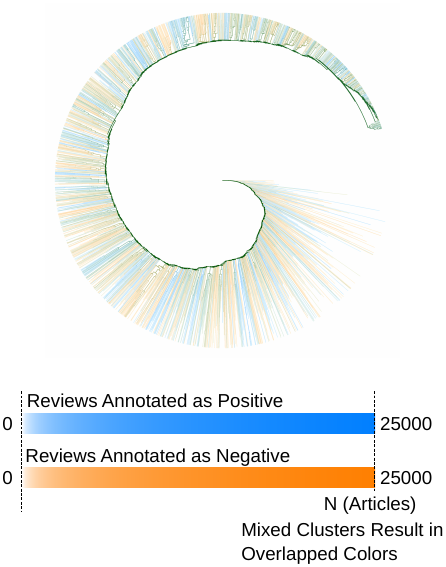}
    \caption{The IMDB 50K Reviews dataset colored by sentiments: The tree was built using the first two principal components derived from SFR-Embedding-Mistral \cite{SFRembedding} embeddings. Parameters yield to the tree building algorithm were: initial $\epsilon = 0.5$, $\Delta\epsilon = 0.0001$, $\mathrm{minpts} = 5$.}
    \label{fig-50k-news-colored}
  \end{center}
\end{figure}

ARI and NMI values remain comparatively low across all tree depths,
with only a modest peaks at layers still in the high dense leaf node range.
This indicates that the embedding models used encode sentiment only weakly relative to other semantic dimensions. The inferred tree, shown in Figure
\ref{fig-50k-news-colored}, reflects this quantitatively observed
behavior: the structure is shallow, with few large subtrees and only
localized regions dominated by either positive or negative reviews. As before,
color intensity follows a logarithmic scale, with white indicating the
absence of reviews of a given label and deeper blue or orange
indicating higher concentration. Overlaps appear in darker,
greenish to black tones. 

Small terminal branches enriched in positive or negative sentiment do
exist, suggesting that repetitive phrasing or stylistic patterns
occasionally induce local clustering. However, no large-scale
sentiment-based separation emerges. These findings highlight an
important methodological point: hierarchical semantic trees faithfully
reflect the information encoded in the embedding space and should not
be expected to recover distinctions that are not strongly represented
by the underlying model.

This observation suggests that the embedding models
used: Qwen3-Embedding-8B \cite{qwen3embedding} and
SFR-Embedding-Mistral \cite{SFRembedding} encode emotional polarity
only to a limited extent. The results highlight the importance of
choosing embedding models appropriate for the expected semantic
structure when constructing trees. 

We also attempted to annotate the IMDB tree with the
Llama-4-Scout-17B-16E-Instruct model \cite{meta2024llama4}. The goal
was to obtain genre-level descriptions of tree regions, under the
assumption that certain movie genres might correlate with
predominantly positive or negative reviews. This attempt was
unsuccessful: the model did not produce reliable or meaningful genre
assignments, and more fundamentally, such information appears to be
largely absent from the review text itself. As such, deriving movie
genres from this dataset seems infeasible with current methods. 

\subsubsection{AG News}

\begin{figure*}[!htb]
  \begin{center}
    \includegraphics[scale=0.8]{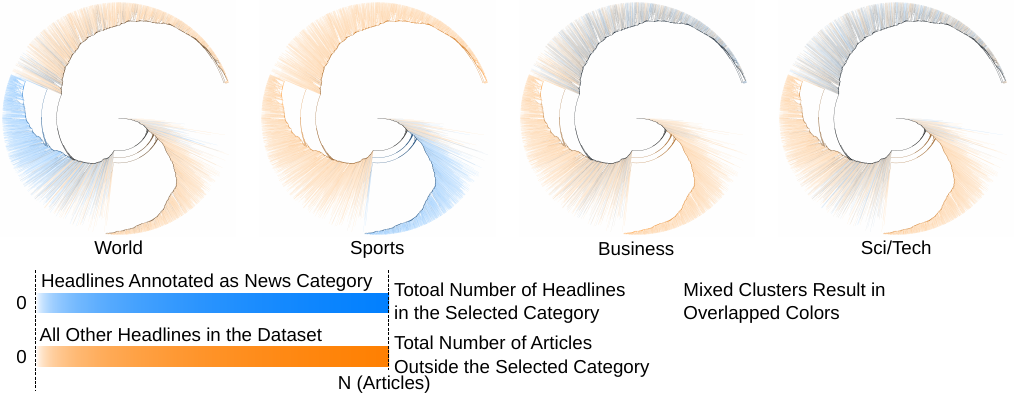}
    \caption{The tree obtained from the AG News dataset in different colorations. The tree shown was inferred from the first two principal components derived from SFR-Embedding-Mistral \cite{SFRembedding} embeddings. Parameters yield to the tree building algorithm were: initial $\epsilon = 0.225$, $\Delta\epsilon = 0.0005$, and $\mathrm{minpts} = 5$}
    \label{fig-ag-news}
  \end{center}
\end{figure*}

\begin{figure}[!htb]
  \begin{center}
    \includegraphics[scale=0.8]{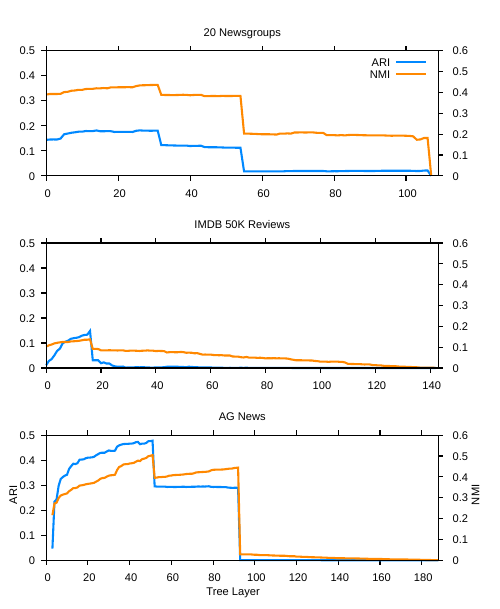}
    \caption{Adjusted Rand Index (ARI) and Normalized Mutual
      Information (NMI) as a function of tree depth for the 20
      Newsgroups, IMDB 50K Reviews and AG News dataset. 20 Newsgroups
      and AG News peak at intermittent clusters indicating that
      semantic agreement with ground-truth labels is maximized at
      intermediate density levels rather than at the finest or
      coarsest partitions.}
    \label{fig-ari-nmi}
  \end{center}
\end{figure}

We further built trees from the AG News dataset and colored them
according to the four available labels: World, Sports, Business, and
Science/Technology. For each label, a separate color map was generated.
Items belonging to the target label are highlighted in blue, while all
other items are shown in orange. As in previous visualizations, we use a
logarithmic color gradient to represent the number of articles in each
cluster, ranging from white (no articles of that type) to fully saturated
colors (all articles in the cluster belong to the target label). The
resulting trees are shown in Figure \ref{fig-ag-news}. 
We observe that the world and sports categories form
clearly separable structures: both branch off early from the root of the
tree, resulting in two well-defined subtrees with only minimal
contamination from other labels. This is also reflected by stronger
ARI and NMI value peaks than in the other datasets as outlined in
Figure \ref{fig-ari-nmi}. Around the root,
however, a small number 
of scattered articles appear, suggesting that certain headlines are not
easily classifiable by the embedding models. These headlines seem to be
very distant from the other news items, only joining the root at a
very diffuse density.

The remaining two categories, business and science/technology, show
substantial overlap. They jointly occupy the third major branch of the
tree, and only small terminal sub-branches demonstrate consistent
label-specific structure. We attribute this to the general type of
embedding models that we are using here and that business headlines
frequently mention science and technology and vice versa in this
dataset.

\subsubsection{The French Theses Dataset: Theses-fr-en}

To further evaluate whether our method can identify and organize broad
and fine-grained disciplinary structure, we applied our tree-building
pipeline to the Theses-fr-en dataset, which consists of
English-language abstracts of French doctoral theses. The dataset
contains author-supplied disciplinary annotations, but these labels
are highly heterogeneous and unstandardized, yielding 13,599 unique
descriptors. Therefore, as outlined in
Section \ref{sec-transcription}, we first performed a cross-annotation
step to obtain more tangible labels. As these labels were generated by
us we did not apply ARI and NMI evaluations on this tree. 

\begin{figure*}[!htb]
  \begin{center}
    \includegraphics[scale=0.8]{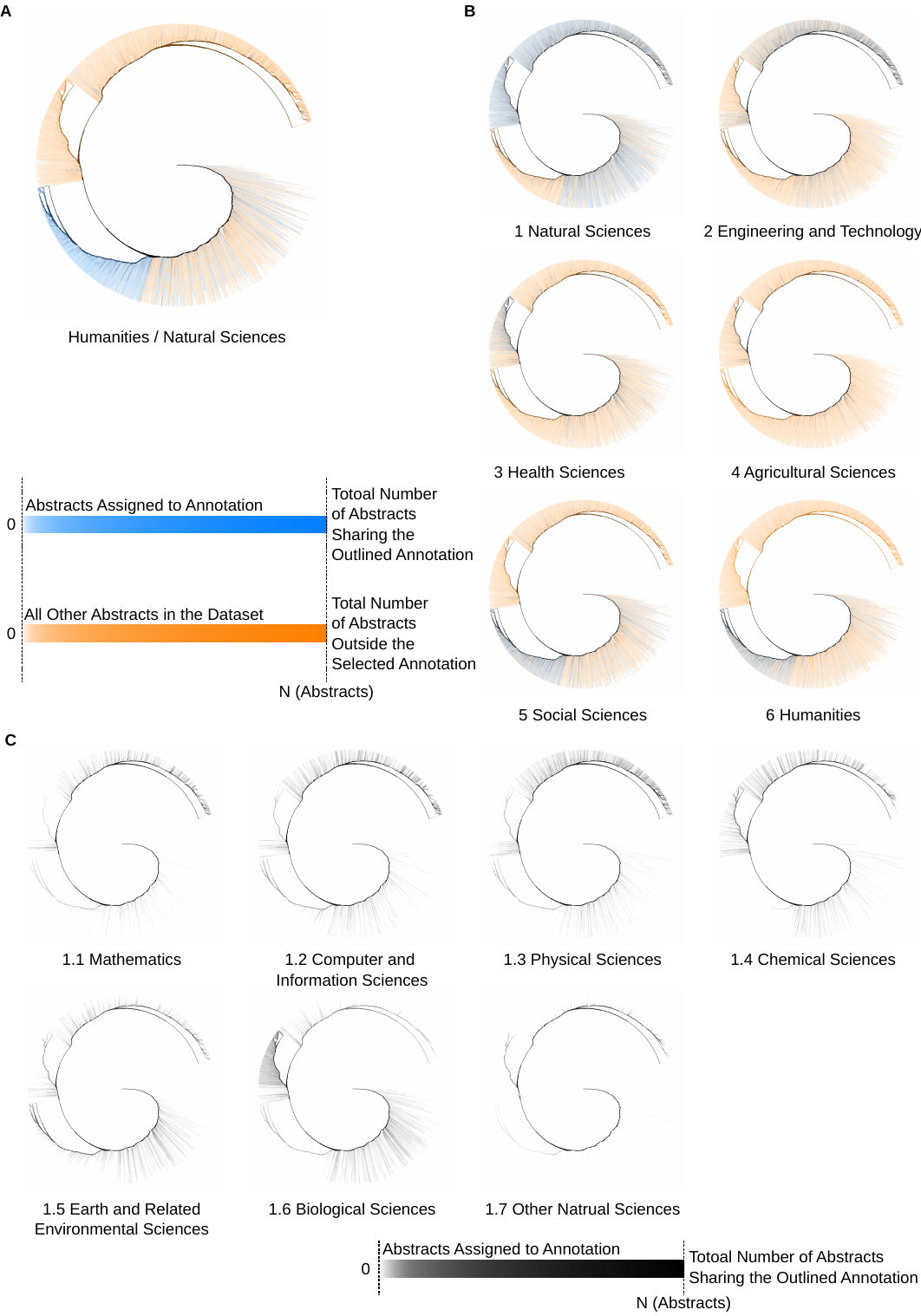}
    \caption{Tree generated from the French Theses theses-fr-en
      dataset using the first two principal components obtained from the embeddings yield by the SFR-Embedding-Mistral \cite{SFRembedding} model:\newline
      \textbf{A}: Coloration according to binary classification Humanities and
      Natural Sciences. \newline
      \textbf{B}: Coloration according to the main groups of the OECD FOS
      classification. \newline
      \textbf{C}: Highlight of the subgroups of the Natural Sciences group
      according to OECD FOS.\newline
      Parameters for tree building were: initial $\epsilon = 0.23$, $\Delta\epsilon = 0.001$, $\mathrm{minpts} = 5$.
    }
    \label{fig-theses}
  \end{center}
\end{figure*}

As an initial experiment, we mapped the original annotations to a
binary classification: humanities versus natural sciences. The
resulting tree clearly separated humanities theses from the remainder
of the corpus, forming a large, coherent branch as shown in Figure
\ref{fig-theses}A. Encouraged by this result, we proceeded to
apply a more detailed transcription based on the OECD Fields of
Science and Technology (FOS) 2002 classification
scheme \cite{oecd-fos}. 

In the tree three major branches are easily identifiable.
The branch identified as “humanities” in the binary test
contains predominantly Social Sciences (5) and Humanities (6)
according to the FOS taxonomy. Social sciences, however, extend beyond
this primary branch, with small isolated scattered clusters appearing
throughout the tree. 

The remaining two major branches contain the Natural Sciences (1), but
with different structural emphases. Strikingly, Engineering and
Technology (2) appears almost exclusively within the very large branch
that occupies nearly the entire upper half of the tree
outlined in Figure \ref{fig-theses}. In contrast, Medical and Health Sciences
(3) predominantly occupy the smaller of the two natural-science
branches, located on the left side of the tree. Examination of FOS
subclasses shows that this medical branch also contains theses
classified as Biological Sciences (1.6), reflecting their conceptual
proximity. 

The pattern of overlaps among FOS categories, particularly the fact
that Natural Sciences (1) span both the engineering-dominant and the
medical-dominant branches, highlights structural ambiguities inherent
in the current FOS hierarchy. Although (1), (2), and (3) are defined
as top-level categories, the tree suggests that Engineering and
Technology (2) and Medical and Health Sciences (3) are more closely
related to different subregions of the Natural Sciences (1) than to
each other. 

Moreover, the tree indicates that (2) and (3) merge into a common
branch earlier than Social Sciences (5) and Humanities (6). After the
social-science and humanities subtree, numerous small and weakly
structured clusters enter the main trunk of the tree (lower right
quadrant), indicating a sizeable number of theses that form islands
in the embedding space and are not well represented by broad
disciplinary groupings. 

All three major branches further subdivide into smaller subtrees
toward their terminal regions. However, we did not find consistent
evidence that these subtrees correspond cleanly to the FOS
subclasses. This lack of correspondence suggests that the tree
structure could support a data-driven reorganizing of disciplinary
taxonomies, potentially capturing conceptual relationships that are
absent, or obscured, in existing classification systems. This
observation directly motivates the subsequent section of the paper. 

\subsubsection{Annotated Institutional Trees: TU Wien and AUB}

\begin{figure*}[!htb]
  \begin{center}
    \includegraphics{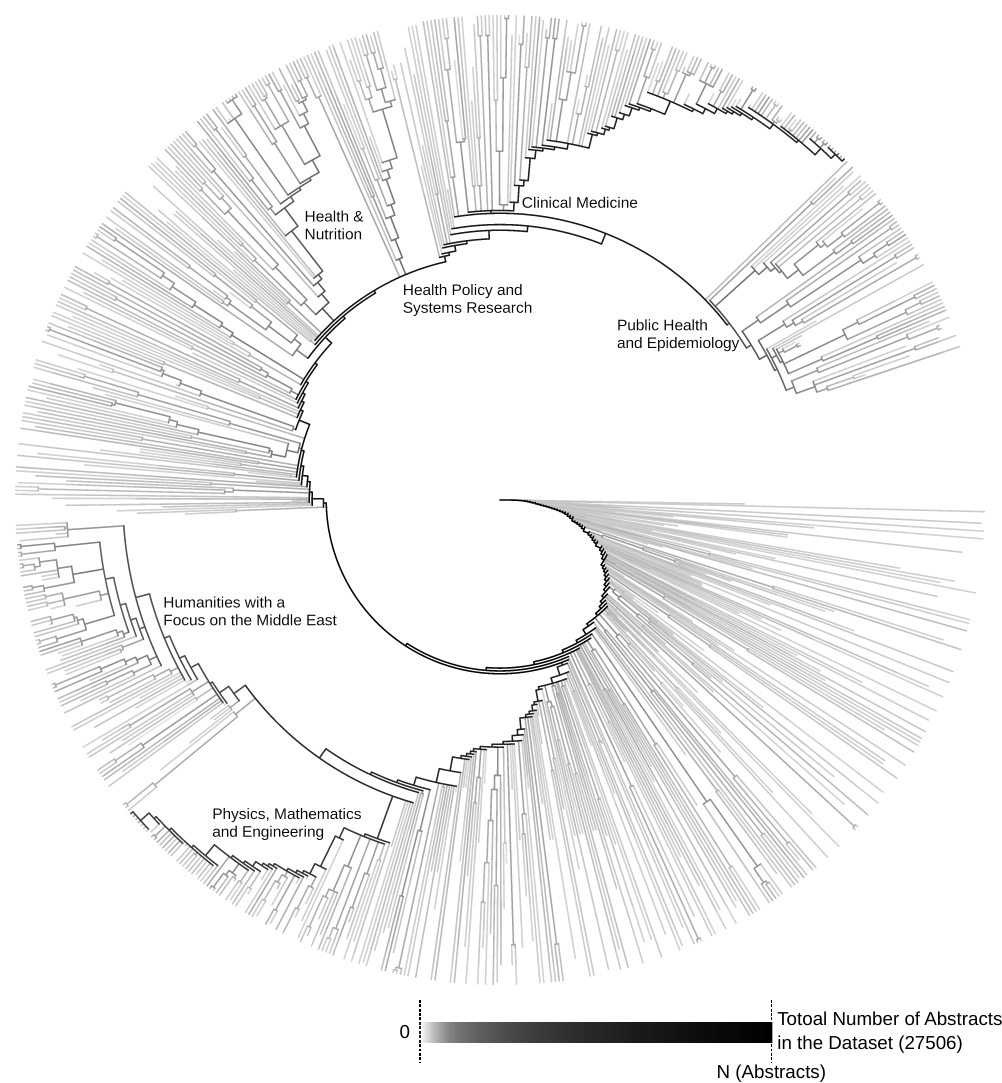}
    \caption{Tree generated from the AUB dataset: The tree shown was inferred from the first two principal components obtained from Qwen3-Embedding-8B
    \cite{qwen3embedding} model embeddings. The tree was built with the following parameters: initial $\epsilon = 0.0045$,
    $\Delta\epsilon = 0.00001$ and $\mathrm{minpts} = 5$.}
    \label{fig-aub-tree}
  \end{center}
\end{figure*}

\begin{figure*}[!htb]
  \begin{center}
    \includegraphics{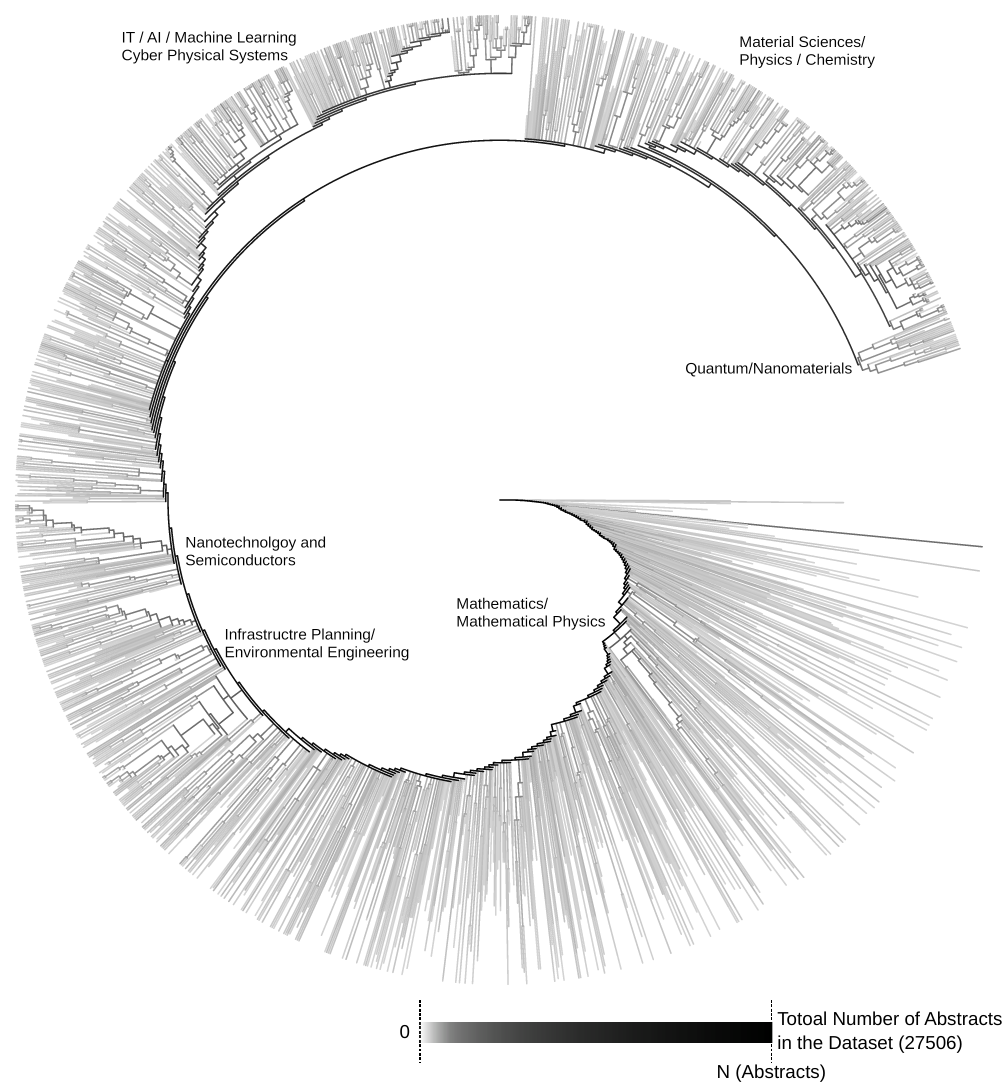}
    \caption{Tree generated from the TU Wien dataset: The tree shown was inferred from the first two principal components obtained from Qwen3-Embedding-8B
    \cite{qwen3embedding} model embeddings. The tree was built with the following parameters: initial $\epsilon = 0.0028$,
    $\Delta\epsilon = 0.000001$ and $\mathrm{minpts} = 5$.}
    \label{fig-tuwien-tree}
  \end{center}
\end{figure*}

Our analysis of the two institutional datasets (TU Wien and AUB) differs
from the benchmark datasets presented earlier. In the benchmark cases,
we projected known labels onto the inferred trees to assess how well
embeddings recover an existing classification. In contrast, the
institutional datasets are used to examine research structures ab
initio: rather than testing whether a predefined taxonomy is recovered,
we investigate how the tree itself organizes the institutional research
landscape. Consequently, ARI and NMI are not
applicable.
Although the embedding models were trained, by their creators, on
external corpora,
and not further fine tuned, the
resulting tree structure is independent of any curated ontology. This
allows us to explore institution-specific disciplinary profiles,
structural proximities between fields, and potential emerging clusters
or cross-disciplinary connections. The results demonstrate that our
method can capture meaningful organizational patterns without relying
on predefined categories.
\begin{itemize}

\item American University of Beirut (AUB)

  At AUB, medicine and health-related
  disciplines dominate the research corpus, reflecting the strong
  presence of the university hospital. This dominance is visible in the
  tree structure, where the entire upper half of the circular tree forms
  the largest and most cohesive subtree. Within this region, several
  well-defined subbranches correspond to specific healthcare domains,
  including health and nutrition, health policy and systems research,
  and public health and epidemiology.

  A distinct humanities subtree is also present, characterized by numerous
  subbranches labeled as Middle Eastern Studies and Conflict
  Resolution. In addition, an engineering subtree that covers engineering,
  physics, and mathematics is clearly visible. Notably, this engineering
  subtree merges with the humanities subtree at a higher density than with the
  medical subtree, suggesting a closer structural relationship between
  engineering and humanities research topics within this corpus than might
  be expected, which might be characteristic of the university's liberal arts tradition.

  To the left of these major subtrees, smaller clusters related to both
  engineering and humanities fields branch into the tree and eventually
  merge with the broader medical subtree. Beyond this point, several
  near-linear terminal branches from a wide range of disciplines connect
  to the tree as outliers before joining the main structure.
  
  The full tree is shown in Figure \ref{fig-aub-tree} and can be explored
  in detail using our interactive tree viewer at \url{https://genealogy.sematlas.com/}. As in the
  other figures, color intensity is logarithmically scaled from white (no
  articles in the branch) to black (all articles in the dataset).
  
\item Technische Universität Wien (TU Wien)

  The tree structure for TU Wien differs markedly from that of AUB. As a
  technical university, TU Wien has a more focused research portfolio,
  which results in a tree that appears more homogeneous overall.
  Nevertheless, the tree reveals several large and meaningful subtrees that
  highlight the institution’s disciplinary strengths.
  
  At the upper right of the tree, we observe a subtree corresponding to
  quantum mechanics, quantum optics, and quantum-scale materials.
  This region merges smoothly into a broad materials science subtree,
  which occupies much of the upper left quadrant of the circular tree.
  
  Also in the upper left quadrant is a large subtree corresponding to
  information technology and computer science, which subdivides into
  numerous smaller branches. Further branches that merge into the main
  tree at later stages correspond to nanotechnology and semiconductor
  engineering. Additional clusters represent infrastructure planning
  and environmental engineering. One of the earliest large branchings
  near the root corresponds to mathematics and theoretical physics,
  forming a distinct foundational science substructure.

  The TU Wien tree is shown in Figure \ref{fig-tuwien-tree}, and an
  interactive exploration is available through our tree viewer at
  \url{https://genealogy.sematlas.com/}. As with the AUB tree, color intensity follows a
  logarithmic scale from white (no articles) to black (all articles in the
  dataset).

\end{itemize}

We note, as can also be verified using our custom tree viewer, that the
automatic annotation of clusters is not without limitations. In many
cases, the LLM assigns overly general labels—such as engineering—to
large or heterogeneous clusters. Moreover, clusters that are
structurally identical across different annotation runs sometimes
receive different labels, suggesting that the model struggles with
processing extremely large input contexts and exhibits instability in
its naming decisions. Despite these limitations, the annotation process
provided substantial insights into both institutions, their publication
profiles, and the major research fields represented within their
corpora. 

\section{Discussion}

Herein we outline our hierarchical density based tree construction
method, applied onto LLM embeddings from large text corpora. Applied on the
abstracts of scientific publications published by the American
University of Beirut and TU Wien, our method produces interpretable,
multi-resolution trees that reveal how topics, semantic fields, and
research areas are organized relative to each other.  By combining
dimensionality reduction, adaptive multi-scale clustering, and
large-context LLM-based annotation, we establish a pipeline capable of
uncovering latent structure even in datasets lacking standardized or
reliable taxonomies.

\subsection{Insights from Benchmark Datasets}

The benchmark datasets provide a reference for the applicability of
our method.

Application of our method on the 20 newsgroups dataset
revealed that the inferred tree was capable to recover intuitive
relationships such as the proximity of sports-related topics,
technical computing groups and overlaps between religious and
political discussions. The hierarchical approach further uncovered that
sports related discussions are further elongated from the different
groups only merging to the whole tree close its root.

The AG news dataset similarly shows that World and Sports articles
form well separated subtrees, but that Business and Science/Technology
headlines exhibit a substantial overlap. This result can intuitively be
explained by the topical similarities between these two
categories. Our tree based method as such creates an intrinsic
semantic organization derived from LLM-embeddings rather than
enforcing a predefined label boundary.

The French doctoral theses dataset provided a further test for our
algorithm. This dataset included ``discipline'' annotations. However
these are rather arbitrarily chosen and unstandardized. We transcribed
these labels first into a binary annotation of either being part of
natural sciences or humanities. To our surprise the theses abstracts
of labeled humanities formed a clear subtree in our tree. Encouraged by
these findings we progressed to transcribe the ``discipline'' labels
into the OECD FOS annotation. We colored the tree according to it and
found thanks to the hierarchical structure that engineering and medical
sciences merge earlier than either social sciences or humanities. A
relationship that as such cannot be derived from the FOS
classification. At the same time we see that finer grained subtrees do
correspond cleanly to the FOS subclasses, indicating that our inferred
structure captures conceptual relationships that differ from
established disciplinary boundaries.

These observations support the idea that our tree-based representations can
serve as a complementary, data-driven alternative to
traditional classification schemes.

The IMDB 50K Reviews dataset yields a very homogeneous tree,
reflecting that the embedding models used in this study
are not able to distinguish the semantic differences between the movie
reviews in a meaningful way. The labels of this dataset are largely
dispersed across the inferred tree, outlining the limitations of our
approach, where the embeddings as outlined here do not encode as here
sentiment or any other required topic separator.   

\subsection{Institution-Specific Research Landscapes}

These datasets are at the core of this research as they illustrate
that using our in here presented pipeline one can derive research
classifications \emph{ab-initio}, under the given that a recent, well
trained LLM-embedding model exists.
In both datasets, from the American University in Beirut and from TU
Wien the inferred trees recover dominant research areas, internal
substructures and relationships between disciplines without relying on
a predefined taxonomy.

At AUB, the strong presence of medical and health-related research
manifests as a large, cohesive subtree branching into public health,
nutrition, epidemiology, and related fields. The emergence of a
well-defined humanities subtree, in particular in Middle Eastern Studies
and Conflict Resolution, highlights the institution’s regional and
cultural focus. The structural proximity between engineering and
humanities observed in the tree suggests interdisciplinary connections
that may not be apparent from administrative classifications alone.

At TU Wien, the tree reflects the institution’s technical orientation,
with prominent subtrees in materials science, quantum physics, information
technology, and engineering. Despite the overall homogeneity of the
corpus, the tree reveals meaningful internal organization.
These findings demonstrate that the method can capture
institution-specific research profiles and internal structure even in
relatively focused corpora.

\subsection{Methodological Considerations}

Several methodological insights emerge from this study. First,
dimensionality reduction via PCA plays a crucial role in enhancing tree
structure. Using $L_2$ distances on a small number of principal
components consistently yielded more articulated and interpretable
trees than cosine distances in the full embedding space. This suggests
that PCA acts as an effective feature-selection mechanism, suppressing
noise and emphasizing the most informative semantic dimensions.

Second, the adaptive multi-$\epsilon$ DBSCAN procedure enables the
recovery of hierarchical structure without imposing a fixed density
threshold. By progressively relaxing density constraints, the method
naturally captures cluster mergers across scales, producing a tree
structure analogous to a dendrogram but grounded in density-based
clustering.

Finally, while LLM-based annotation substantially improves
interpretability, it also introduces limitations. Large clusters often
receive overly general or unstable labels, reflecting current
constraints of long-context inference. However, even imperfect
annotations provide valuable guidance for navigating and interpreting
complex tree structures.

\subsection{Limitations and Future Directions}

Inferred trees are inherently constrained by the representational
capacity of the embedding models. Semantic dimensions that are weakly
encoded, such as fine-grained sentiment or subtle disciplinary
distinctions, cannot be reliably recovered through clustering alone.
Future work could explore domain-specific or task-adapted embedding
models to address this limitation.

In addition, annotation remains computationally expensive and sensitive
to model variability. More robust strategies, such as hierarchical
annotation, or summarization strategies tied together with annotation
strategies may improve stability and scalability.

Looking beyond text-based corpora, a particularly promising direction
is the extension of this framework to other data modalities. The
method itself is agnostic to the origin of the embeddings and could, in
principle, be applied to image embeddings, audio embeddings, or
multimodal representations. For example, trees inferred from image
embeddings could reveal visual themes or stylistic groupings, while
trees built from music embeddings might uncover genre structure,
rhythmic patterns, or harmonic similarities. Applying the proposed
approach to such modalities would allow for the exploration of latent
structure in domains where hierarchical organization is often implicit
rather than explicitly labeled.

\section{Conclusion}
Overall, our results indicate that hierarchical density clustering of
LLM embeddings provides a powerful and flexible framework for mapping
semantic structure in large text corpora. The method
reveals meaningful relationships that both complement and challenge
existing classification systems. By enabling data-driven,
multi-resolution representations of textual domains, this approach
offers new opportunities for research assessment, knowledge mapping,
and other exploratory analysis.

\section*{Acknowledgments}
TH thanks the TU Wien Campus IT - Datalab - AI Team and EuroCC Austria for supporting this research.  
JB thanks the Center for Advanced Mathematical Sciences (CAMS) and the Artificial Intelligence, Data Science and Computing Hub (AI-DSC), for supporting this research.

We also thank the TU Wien Service Unit for Research Information Systems for providing the TU Wien dataset, the library of the American University of Beirut for providing the AUB dataset and Austrian Scientific Computing (ASC) for granting early access to the MUSICA supercomputing infrastructure: \newline \url{https://asc.ac.at}

\section*{Code and Data Availability}
Code, TU Wien and AUB Datasets are available on github: \newline \url{https://github.com/haschka/semantic-trees}

\bibliographystyle{ieeetr}
\bibliography{bib}

@misc{embeddings-proof,
      title={Language-agnostic BERT Sentence Embedding}, 
      author={Fangxiaoyu Feng and Yinfei Yang and Daniel Cer and Naveen Arivazhagan and Wei Wang},
      year={2022},
      eprint={2007.01852},
      archivePrefix={arXiv},
      primaryClass={cs.CL},
 doi={10.48550/arXiv.2007.01852}
}

@misc{embeddings-proof-more,
      title={Evaluating NLP Embedding Models for Handling Science-Specific Symbolic Expressions in Student Texts}, 
      author={Tom Bleckmann and Paul Tschisgale},
      year={2025},
      eprint={2505.17950},
      archivePrefix={arXiv},
      primaryClass={cs.CL},
      doi={10.48550/arXiv.2505.17950} 
}

@article{embedding-proof-even-more,
title = {Embedding Models: A Comprehensive Review with Task-Oriented Assessment},
journal = {International Journal of Advanced Computer Science and Applications},
doi = {10.14569/IJACSA.2025.0161056},
year = {2025},
publisher = {The Science and Information Organization},
volume = {16},
number = {10},
author = {Lahbib Ajallouda and Meriem Hassani Saissi and Ahmed Zellou}
}

@article{bagofwords,
author = {Zellig S. Harris},
title = {Distributional Structure},
journal = {WORD},
volume = {10},
number = {2-3},
pages = {146--162},
year = {1954},
publisher = {Routledge},
doi = {10.1080/00437956.1954.11659520}
}

@article{tfidf,
    author = {Spark Jones, Karen},
    title = {A Statistical Interpretation of Term Specificity and its Application in Retrieval},
    journal = {Journal of Documentation},
    volume = {28},
    number = {1},
    pages = {11-21},
    year = {1972},
    month = {01},
    issn = {0022-0418},
    doi = {10.1108/eb026526}
}

@article{lsa,
author="Foltz, Peter W.",
title="Latent semantic analysis for text-based research",
journal="Behavior Research Methods, Instruments, {\&} Computers",
year="1996",
month="Jun",
day="01",
volume="28",
number="2",
pages="197--202",
issn="1532-5970",
doi="10.3758/BF03204765"
}

@article{lsa-first,
author = {Deerwester, Scott and Dumais, Susan T. and Furnas, George W. and Landauer, Thomas K. and Harshman, Richard},
title = {Indexing by latent semantic analysis},
journal = {Journal of the American Society for Information Science},
volume = {41},
number = {6},
pages = {391-407},
doi = {10.1002/(SICI)1097-4571(199009)41:6<391::AID-ASI1>3.0.CO;2-9},
year = {1990}
}

@article{lsa-intro,
author = {Thomas K Landauer and Peter W. Foltz and Darrell Laham},
title = {An introduction to latent semantic analysis},
journal = {Discourse Processes},
volume = {25},
number = {2-3},
pages = {259--284},
year = {1998},
publisher = {Routledge},
doi = {10.1080/01638539809545028},
}

@article{lda,
author = {Blei, David M. and Ng, Andrew Y. and Jordan, Michael I.},
title = {Latent dirichlet allocation},
year = {2003},
issue_date = {3/1/2003},
publisher = {JMLR.org},
volume = {3},
number = {null},
issn = {1532-4435},
journal = {J. Mach. Learn. Res.},
month = mar,
pages = {993–1022},
numpages = {30},
doi = {10.5555/944919.944937}
}

@article{hda,
author = {Yee Whye Teh and Michael I Jordan and Matthew J Beal and David M Blei},
title = {Hierarchical Dirichlet Processes},
journal = {Journal of the American Statistical Association},
volume = {101},
number = {476},
pages = {1566--1581},
year = {2006},
publisher = {ASA Website},
doi = {10.1198/016214506000000302}
}

@inproceedings{interpretability,
author = {Chang, Jonathan and Boyd-Graber, Jordan and Gerrish, Sean and Wang, Chong and Blei, David M.},
title = {Reading tea leaves: how humans interpret topic models},
year = {2009},
isbn = {9781615679119},
publisher = {Curran Associates Inc.},
address = {Red Hook, NY, USA},
booktitle = {Proceedings of the 23rd International Conference on Neural Information Processing Systems},
pages = {288–296},
numpages = {9},
location = {Vancouver, British Columbia, Canada},
series = {NIPS'09}
}

@misc{word2vec,
      title={Efficient Estimation of Word Representations in Vector Space}, 
      author={Tomas Mikolov and Kai Chen and Greg Corrado and Jeffrey Dean},
      year={2013},
      eprint={1301.3781},
      archivePrefix={arXiv},
      primaryClass={cs.CL},
      doi={10.48550/arXiv.1301.3781}
}

@inproceedings{transformer,
author = {Vaswani, Ashish and Shazeer, Noam and Parmar, Niki and Uszkoreit, Jakob and Jones, Llion and Gomez, Aidan N. and Kaiser, \L{}ukasz and Polosukhin, Illia},
title = {Attention is all you need},
year = {2017},
isbn = {9781510860964},
publisher = {Curran Associates Inc.},
address = {Red Hook, NY, USA},
booktitle = {Proceedings of the 31st International Conference on Neural Information Processing Systems},
pages = {6000–6010},
numpages = {11},
location = {Long Beach, California, USA},
doi = {10.5555/3295222.3295349},
series = {NIPS'17}
}

@misc{rag,
      title={Retrieval-Augmented Generation for Knowledge-Intensive NLP Tasks}, 
      author={Patrick Lewis and Ethan Perez and Aleksandra Piktus and Fabio Petroni and Vladimir Karpukhin and Naman Goyal and Heinrich Küttler and Mike Lewis and Wen-tau Yih and Tim Rocktäschel and Sebastian Riedel and Douwe Kiela},
      year={2021},
      eprint={2005.11401},
      archivePrefix={arXiv},
      primaryClass={cs.CL},
      doi={10.48550/arXiv.2005.11401}
}

@misc{llm-embed-rag,
      title={Each to Their Own: Exploring the Optimal Embedding in RAG}, 
      author={Shiting Chen and Zijian Zhao and Jinsong Chen},
      year={2025},
      eprint={2507.17442},
      archivePrefix={arXiv},
      primaryClass={cs.CL},
      doi={10.48550/arXiv.2507.17442}
}

@article{k-means-first,
  title={Some Methods for Classification and Analysis of Multivariate Observations},
  author={MacQueen, James},
  booktitle={Proceedings of the Fifth Berkeley Symposium on Mathematical Statistics and Probability},
  year={1967}
}

@article{k-means,
title = {Data clustering: 50 years beyond K-means},
journal = {Pattern Recognition Letters},
volume = {31},
number = {8},
pages = {651-666},
year = {2010},
note = {Award winning papers from the 19th International Conference on Pattern Recognition (ICPR)},
issn = {0167-8655},
doi = {https://doi.org/10.1016/j.patrec.2009.09.011},
url = {https://www.sciencedirect.com/science/article/pii/S0167865509002323},
author = {Anil K. Jain}
}

@article{hierarchical-first,
author={Johnson, Stephen C.},
title={Hierarchical clustering schemes},
journal={Psychometrika},
year={1967},
month={Sep},
day={01},
volume={32},
number={3},
pages={241--254},
issn={1860-0980},
doi={10.1007/BF02289588}
}

@misc{hierarchical,
      title={Modern hierarchical, agglomerative clustering algorithms}, 
      author={Daniel M\"ullner},
      year={2011},
      eprint={1109.2378},
      archivePrefix={arXiv},
      primaryClass={stat.ML},
      doi={10.48550/arXiv.1109.2378}
}

@article{hdbscan,
doi = {10.21105/joss.00205},
url = {https://doi.org/10.21105/joss.00205},
year = {2017},
publisher = {The Open Journal},
volume = {2},
number = {11},
pages = {205},
author = {McInnes, Leland and Healy, John and Astels, Steve},
title = {hdbscan: Hierarchical density based clustering},
journal = {Journal of Open Source Software} }

@article{mnhn-tree-tools,
    author = {Haschka, Thomas and Ponger, Loic and Escud{\'{e}}, Christophe and Mozziconacci, Julien},
    title = {MNHN-Tree-Tools: a toolbox for tree inference using multi-scale clustering of a set of sequences},
    journal = {Bioinformatics},
    volume = {37},
    number = {21},
    pages = {3947-3949},
    year = {2021},
    month = {06},
    issn = {1367-4803},
    doi = {10.1093/bioinformatics/btab430},
}

@ARTICLE{md-tree-tools,
AUTHOR={Haschka, Thomas  and Lamari, Foudil  and Mochel, Fanny  and Zujovic, Violetta },
TITLE={Innovative tree-based method for sampling molecular conformations: exploring the ATP-binding cassette subfamily D member 1 (ABCD1) transporter as a case study},
JOURNAL={Frontiers in Molecular Biosciences},
VOLUME={Volume 11 - 2024},
YEAR={2024},
DOI={10.3389/fmolb.2024.1440529},
ISSN={2296-889X}
}

@inproceedings{dbscan,
author = {Ester, Martin and Kriegel, Hans-Peter and Sander, J\"{o}rg and Xu, Xiaowei},
title = {A density-based algorithm for discovering clusters in large spatial databases with noise},
year = {1996},
publisher = {AAAI Press},
booktitle = {Proceedings of the Second International Conference on Knowledge Discovery and Data Mining},
pages = {226–231},
numpages = {6},
location = {Portland, Oregon},
series = {KDD'96},
doi = {10.5555/3001460.3001507}
}

@article{scikit-learn,
  title={Scikit-learn: Machine Learning in {P}ython},
  author={Pedregosa, F. and Varoquaux, G. and Gramfort, A. and Michel, V.
          and Thirion, B. and Grisel, O. and Blondel, M. and Prettenhofer, P.
          and Weiss, R. and Dubourg, V. and Vanderplas, J. and Passos, A. and
          Cournapeau, D. and Brucher, M. and Perrot, M. and Duchesnay, E.},
  journal={Journal of Machine Learning Research},
  volume={12},
  pages={2825--2830},
  year={2011}
}

@misc{20newsgroups,
  author       = {Mitchell, Tom},
  title        = {{Twenty Newsgroups}},
  year         = {1997},
  howpublished = {UCI Machine Learning Repository},
  doi = {10.24432/C5C323}
}

@InProceedings{50kmovies,
author = {Maas, Andrew L. and Daly, Raymond E. and Pham, Peter T. and Huang, Dan and Ng, Andrew Y. and Potts, Christopher},
title = {Learning Word Vectors for Sentiment Analysis},
booktitle = {Proceedings of the 49th Annual Meeting of the Association for Computational Linguistics: Human Language Technologies},
month = {June},
year = {2011},
address = {Portland, Oregon, USA},
publisher = {Association for Computational Linguistics},
pages = {142--150},
url = {http://www.aclweb.org/anthology/P11-1015} }

@inproceedings{agnews,
author = {Zhang, Xiang and Zhao, Junbo and LeCun, Yann},
title = {Character-level convolutional networks for text classification},
year = {2015},
publisher = {MIT Press},
address = {Cambridge, MA, USA},
booktitle = {Proceedings of the 29th International Conference on Neural Information Processing Systems - Volume 1},
pages = {649–657},
numpages = {9},
location = {Montreal, Canada},
series = {NIPS'15},
doi = {10.5555/2969239.2969312}
}

@misc{qwen3embedding,
title={Qwen3 Embedding: Advancing Text Embedding and Reranking Through Foundation Models},
author={Yanzhao Zhang and Mingxin Li and Dingkun Long and Xin Zhang and Huan Lin and Baosong Yang and Pengjun Xie and An Yang and Dayiheng Liu and Junyang Lin and Fei Huang and Jingren Zhou},
year={2025},
archivePrefix={arXiv},
primaryClass={cs.CL},
doi={10.48550/arXiv.2506.05176},
}

@misc{SFRembedding,
  title={SFR-Embedding-Mistral:Enhance Text Retrieval with Transfer Learning},
  author={Rui Meng and Ye Liu and Shafiq Rayhan Joty and Caiming Xiong and Yingbo Zhou and Semih Yavuz},
  howpublished={Salesforce AI Research Blog},
  year={2024},
  url={https://www.salesforce.com/blog/sfr-embedding/}
}

@misc{meta2024llama4,
title={Introducing LLaMA 4: Advancing Multimodal Intelligence},
author={Meta AI},
year={2024},
url={https://ai.meta.com/blog/llama-4-multimodal-intelligence/} }

@misc{gpt-oss,
      title={gpt-oss-120b \& gpt-oss-20b Model Card}, 
      author={OpenAI and Sandhini Agarwal and Lama Ahmad and Jason Ai and Sam Altman and Andy Applebaum and Edwin Arbus and Rahul K. Arora and Yu Bai and Bowen Baker and Haiming Bao and Boaz Barak and Ally Bennett and Tyler Bertao and Nivedita Brett and Eugene Brevdo and Greg Brockman and Sebastien Bubeck and Che Chang and Kai Chen and Mark Chen and Enoch Cheung and Aidan Clark and Dan Cook and Marat Dukhan and Casey Dvorak and Kevin Fives and Vlad Fomenko and Timur Garipov and Kristian Georgiev and Mia Glaese and Tarun Gogineni and Adam Goucher and Lukas Gross and Katia Gil Guzman and John Hallman and Jackie Hehir and Johannes Heidecke and Alec Helyar and Haitang Hu and Romain Huet and Jacob Huh and Saachi Jain and Zach Johnson and Chris Koch and Irina Kofman and Dominik Kundel and Jason Kwon and Volodymyr Kyrylov and Elaine Ya Le and Guillaume Leclerc and James Park Lennon and Scott Lessans and Mario Lezcano-Casado and Yuanzhi Li and Zhuohan Li and Ji Lin and Jordan Liss and Lily and Liu and Jiancheng Liu and Kevin Lu and Chris Lu and Zoran Martinovic and Lindsay McCallum and Josh McGrath and Scott McKinney and Aidan McLaughlin and Song Mei and Steve Mostovoy and Tong Mu and Gideon Myles and Alexander Neitz and Alex Nichol and Jakub Pachocki and Alex Paino and Dana Palmie and Ashley Pantuliano and Giambattista Parascandolo and Jongsoo Park and Leher Pathak and Carolina Paz and Ludovic Peran and Dmitry Pimenov and Michelle Pokrass and Elizabeth Proehl and Huida Qiu and Gaby Raila and Filippo Raso and Hongyu Ren and Kimmy Richardson and David Robinson and Bob Rotsted and Hadi Salman and Suvansh Sanjeev and Max Schwarzer and D. Sculley and Harshit Sikchi and Kendal Simon and Karan Singhal and Yang Song and Dane Stuckey and Zhiqing Sun and Philippe Tillet and Sam Toizer and Foivos Tsimpourlas and Nikhil Vyas and Eric Wallace and Xin Wang and Miles Wang and Olivia Watkins and Kevin Weil and Amy Wendling and Kevin Whinnery and Cedric Whitney and Hannah Wong and Lin Yang and Yu Yang and Michihiro Yasunaga and Kristen Ying and Wojciech Zaremba and Wenting Zhan and Cyril Zhang and Brian Zhang and Eddie Zhang and Shengjia Zhao},
      year={2025},
      eprint={2508.10925},
      archivePrefix={arXiv},
      primaryClass={cs.CL},
      doi={10.48550/arXiv.2508.10925}
}

@misc{llama-cpp,
title={Introduction to ggml},
author={Xuan-Son Nguyen and Georgi Gerganov and S. Laren},
year={2024},
url={https://huggingface.co/blog/introduction-to-ggml}
}

@book{oecd-fos,
  author    = "{OECD}",
  title     = "{The Measurement of Scientific and Technological Activities: Frascati Manual 2002 — Proposed Standard Practice for Surveys on Research and Experimental Development}",
  publisher = "{OECD Publishing}",
  address   = "{Paris, France}",
  year      = {2002},
  isbn      = {9789264199040},
  note      = {Includes the “Fields of Science (FOS)” classification (2002) for R\&D statistics},
  doi       = {10.1787/9789264199040-en}
}

@article{newick-utilities,
    author = {Junier, Thomas and Zdobnov, Evgeny M.},
    title = {The Newick utilities: high-throughput phylogenetic tree processing in the Unix shell},
    journal = {Bioinformatics},
    volume = {26},
    number = {13},
    pages = {1669-1670},
    year = {2010},
    month = {05},
    issn = {1367-4803},
    doi = {10.1093/bioinformatics/btq243},
}

\end{document}